\documentclass{bmvc2k}
\usepackage{amssymb}
\usepackage{amsmath}
\usepackage{enumitem}
\usepackage{multirow}
\usepackage{url}
\usepackage{wrapfig,lipsum,booktabs}

\usepackage[title]{appendix}
\usepackage{multirow}

\title{GestSync: Determining who is speaking without a talking head}

\addauthor{Sindhu B Hegde}{sindhu@robots.ox.ac.uk}{1}
\addauthor{Andrew Zisserman}{az@robots.ox.ac.uk}{1}

\addinstitution{
 Visual Geometry Group\\
 Department of Engineering Science\\
 University of Oxford\\
 Oxford, UK
}

\runninghead{Hegde, Zisserman}{GestSync}

\begin{document}

\maketitle

\begin{abstract}
In this paper we introduce a new synchronisation task, {\em Gesture-Sync}: determining if a person's gestures are correlated with their speech or not. In comparison to Lip-Sync, Gesture-Sync is far more challenging as there is a far looser relationship between the voice and body movement than there is between voice and lip motion. We introduce a dual-encoder model for this task, and compare a number of input representations including RGB frames, keypoint images, and keypoint vectors, assessing their performance and advantages. We show that the model can be trained using self-supervised learning alone, and evaluate its performance on the LRS3 dataset. Finally, we demonstrate applications of Gesture-Sync for audio-visual synchronisation, and in determining who is the speaker in a crowd, without seeing their faces. The code, datasets and pre-trained models can be found at: \url{https://www.robots.ox.ac.uk/~vgg/research/gestsync}.
\end{abstract}

\vspace{-8pt}
\begin{figure}[ht]
\includegraphics[width=\textwidth]{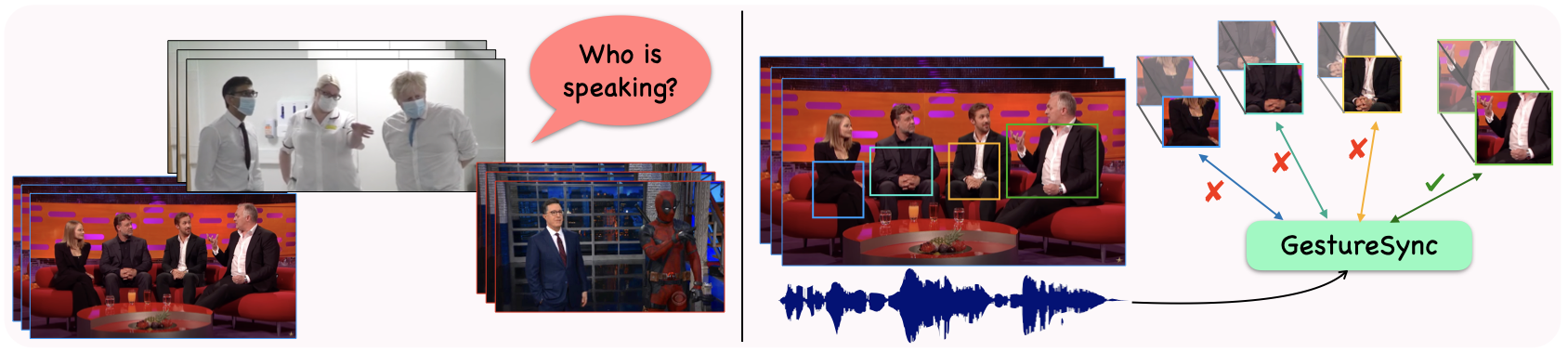}
  \caption{Who is speaking in these scenes?
Our model, dubbed {\em GestSync} learns to identify whether a person's gestures and speech are ``in-sync''. The learned embeddings from our model are used to determine ``who is speaking'' in the crowd, without looking at their faces. Please refer to the demo video for examples.}
  \label{fig:teaser}
  \vspace{-15pt}
\end{figure}

\section{Introduction}
\label{sec:intro}

Imagine you are watching a movie scene where multiple people are in a conversation, but the camera view does not include their faces; how can you determine who is speaking? That it is possible at all is due to the motion of the speaker's hands and arms, that may be correlated with the speech, as illustrated in Figure~\ref{fig:teaser}.

If the face is visible, then the synchronisation of the lip motion and speech provides a very strong signal for who is speaking. This {\em lip-sync} problem has been well explored and solved in prior work~\cite{Chung16a,Chung2019perfect,Afouras20b,halperin2019dynamic,Khosravan19,kim2021end,kadandale2022vocalist}. It is a strong
signal because speech and lip motion are dense in time, and causally related. However, there could be various scenarios where facial features, especially lips, may not be accessible or useful. Examples include: (i) detecting if a person is speaking from a long distance where the lips are too small to decipher their movements; (ii) people speaking into a microphone that occludes their lips; (iii) scenes in movies where there are often lots of pose and scene changes along with occlusion of face by other actors/objects; and (iv) ego-centric views of the wearer's hands from a wearable device. In all these cases, relying on traditional lip synchronisation signals will not be applicable. Thus, our objective in this paper is to determine to what extent it is possible to answer the question `who is speaking?' {\em from gestures alone}. This is very challenging because the correlation between speech and gestures is much looser than that between speech and lip motion.  

McNeill~\cite{Mcneill94} categorizes co-speech gestures into four types: {\em beats}, {\em iconics}, {\em metaphorics}, and {\em deictics}. Beat gestures are usually two phase hand movements (up/down, left/right etc) that mark speech prosody, reinforce speech intonation, or control speech flow. They do not carry semantic content~\cite{Andric12} but are used to emphasize particular words or phrases, and match the rhythm of the speech. Iconics physically represent the semantic content of speech, e.g.\ an arch-like hand movement accompanying the word ``bridge''.  Metaphoric gestures represent abstract ideas that have no material instantiation or have metaphorical meaning -- e.g.\  an upward gesture accompanying ``the stock price went through the roof''.  Deictics are pointing gestures that indicate points or directions of interest. Note, these spontaneous gestures are very different from the structured gestures used in sign language~\cite{sutton-spence_woll_1999} or typical gesture recognition datasets, e.g.~\cite{Wan16chalearn}, and furthermore they are very  speaker dependent -- some speakers make little or no gestures when they speak.

We introduce a dual-encoder model for this task, where the visual and audio streams are ingested by different encoders and then compared. However, rather than directly using RGB frames, we first investigate the use of $2$D pose keypoints, such as points marking out the lips, elbows or shoulders. There are three reasons for this: first, we can effortlessly `switch-off' information by not including a subset of keypoints. So, for example, if no facial keypoints are included, then the model cannot make use of lip-sync cues at all for the task. Second, keypoints remove many `nuisance' parameters for the task, such as lighting and clothing, and are a much more compressed representation than images,  so the
training and exploration of models is easier. Finally, keypoints can be computed at frame rate, so models for the task can run in real time. We also investigate keypoints for the task of lip-sync as this enables performance to be compared with previous work on lip-sync -- since there is no prior work on gesture-sync that we know of to compare with.

For training, we follow the example of lip-sync, where models are trained using self-supervised learning by introducing temporal offsets, and are then used for the task of active-speaker detection. We demonstrate in this paper that {\em gesture-sync} models can also be trained by a similar method of self-supervision. However, it is necessary to introduce longer temporal offsets, as hand motions correlated with speech are slower than lip motions, and so a longer temporal scale must be used.

In summary, we make the following four contributions: (i) we introduce a new task of audio-visual synchronisation using gestures; (ii) we propose a model for this task, and show that it can be trained using self-supervised learning; (iii) we investigate different representations of the video data including keypoints, RGB frames, and keypoints rendered as images;  finally, (iv), we show very promising quantitative results for this task on the LRS3 dataset~\cite{LRS3_2018_Afouras}, and outline a number of applications.

\section{Related Work}

\noindent
\textbf{Lip synchronisation:} Before deep learning, lip-speech synchronisation was predicted using hand-crafted features, e.g.\ \cite{slaney2000facesync}. In the deep learning era, SyncNet, a dual encoder with one ConvNet for the the visual stream and another for the audio stream, was trained using self-supervised learning~\cite{Chung16a}. The model architecture and training were then improved over a series of papers~\cite{Chung2019perfect,halperin2019dynamic,kim2021end,kadandale2022vocalist}. All of these required a face detector. More recent work~\cite{Khosravan19} has demonstrated that with the introduction of attention mechanisms, synchronisation for talking heads could be learnt at the image level and   also the talking head could be tracked~\cite{Afouras20b}.\\

\noindent
\textbf{Synchronisation from sparse signals:}
While lip movements are densely correlated with audio in time, other audio-visual associations are more sparse. \cite{Chen21b,Iashin22} explore audio-visual synchronisation for arbitrary videos in the wild. For example, in a clip of a dog, the audio and video signals only provide a synchronisation signal when the dog barks and not at other time. Our work explores a task with a similar nature. The gestures match with the speech quite clearly at certain time-steps, but could be quite ambiguous at others. Long-term context becomes crucial here in order to confidently predict an audio-visual match.\\

\noindent
\textbf{Keypoint representations:}
While early methods relied on handcrafted features, recent approaches leverage deep learning architectures like CNNs to estimate accurate human body keypoints~\cite{CNNPoseEst_2016, DensePose_2018_CVPR, OpenPose_2019_Cao, Pfister15a, Mediapipe_2019_Lugaresi}. Several works have considered using keypoints as inputs for various tasks. One of the popular ways is to construct graph neural networks (GNNs) using keypoint representations. Tasks such as action recognition~\cite{STGCNActionRecog_2018}, gesture recognition~\cite{GNNGestureRecog_2021}, and sign language segmentation and recognition~\cite{STGCNSignLangRecog_2019, AutomaticSignSeg_2020_ECCV, IsolatedSignRecog_2021, AttnSignRecog_2021} have been attempted using GNNs that take in keypoints as inputs. These works have shown impressive results, indicating the effectiveness of using keypoint representations. In our work, we take a slightly different path: directly using keypoint representations as inputs to a Transformer based model. \\

\noindent
\textbf{Speech and gestures:} \cite{Speech2Gesture_2019_CVPR}  generates plausible hand/arm gesture sequences from a speech signal. A GAN based network is employed to perform cross-modal translation from speech to hand and arm motion. The paper focuses on generating gestures by training on each speaker separately. In this paper, we explore a different task of learning synchronisation between speech and gestures while also training on a large number of identities in the wild.

\vspace{-10pt}
\section{Learning to Synchronise Gestures and Speech}
\label{sec:arch}

\begin{figure}[h]
  \centering
\includegraphics[width=\linewidth]{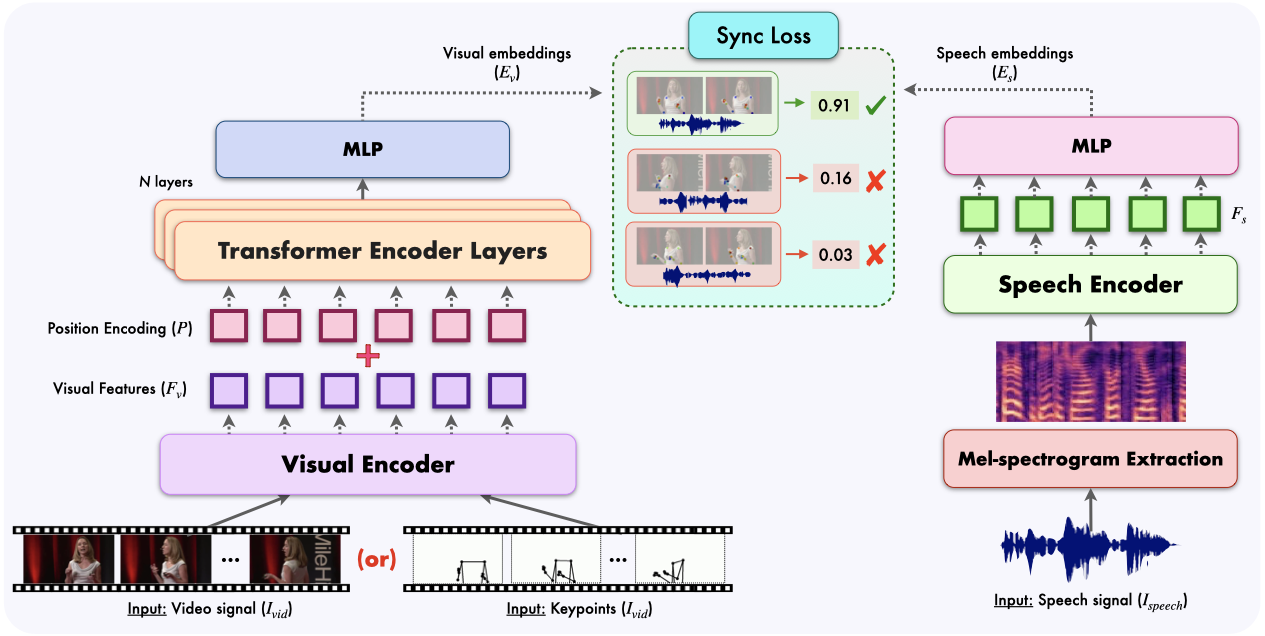}
  \vspace{-10pt}
  \caption{\textbf{The GestSync Network:} The model ingests one of the following video representations: (i) raw RGB frames, (ii) keypoint skeleton images, or (iii) keypoint vectors. The visual encoder processes the input signal and generates the visual features. These features are processed by the Transformer model to obtain the visual embeddings. For the speech input, the speech encoder uses Mel-spectrogram representation and generates the speech embeddings. The model is trained with a contrastive loss to have a high similarity between the visual and speech embeddings when synchronized, and low otherwise.}
  \label{fig:arch}
  \vspace{-10pt}
\end{figure}

Figure~\ref{fig:arch} shows an overview of the proposed {\em GestSync} network. The network takes in a video sequence, denoted as $I_{vid} = \{ v_1, v_2, ..., v_m \}$, and a
speech sequence, denoted as $I_{speech} = \{ s_1, s_2, ..., s_n
\}$.  The video and audio streams are processed independently to obtain two distinct embeddings: visual embedding $E_{v}$ and speech embedding $E_{s}$. Our goal is to determine the
correlation between gestures and speech patterns, which we approach as a synchronisation task. To accomplish this, we compute the similarity between the visual and speech embeddings. Below, we provide a detailed description of the modules involved, followed by the training procedure in
Section~\ref{sec:training}, and implementation details in Section~\ref{sec:implem}.

\vspace{-10pt}
\subsection{Video Representations}
We consider multiple
input video representations: (i) \textbf{RGB frames:} Raw RGB frames are used as  input; (ii) \textbf{Keypoint images:} The extracted keypoints are rendered onto an  image to create a skeleton graph using the  keypoint $(x,y)$ coordinates; (iii) \textbf{Keypoint vectors:} The keypoints are used directly as vectors. Figure~\ref{fig:arch} illustrates the RGB image representation and keypoint skeleton image representations. \\

\noindent
\textbf{Visual Encoder:} 
We extract the visual features using a convolutional neural network (CNN). The input to the CNN is either the RGB frames or keypoint skeleton images or keypoint vectors. The aim (for both image and vector inputs) is to obtain the visual embeddings which we utilize further in computing the audio-visual similarity. For RGB frame and keypoint skeleton image inputs,
we use $3$D convolution layers, similar to previous audio-visual networks~\cite{Chung2019perfect, Afouras20b}, where the first layer has a temporal receptive field of $5$ frames to capture the motion information. The network processes $(T\times 3\times H\times W)$-dimension frames (where $T$ is the temporal window, and $H$ and $W$ are the frame height and width respectively), and generates visual features $F_v$ as $T$ vectors of dimension $d$. For keypoint vector inputs, we use $2$D CNNs. The model takes in data
of dimension $(T\times 2\times N_{kp})$ (where $N_{kp}$ is the number of keypoints), and these are treated as images of height $\times$ width $(T\times  N_{kp})$ with the $(x,y)$ coordinate dimension acting as two channels. The output visual features $F_v$ are
$T$ vectors of dimension $d\times N_{kp}$.\\

\noindent
\textbf{Transformer Encoder:}
The visual features $F_v$ are encoded using a Transformer Encoder with $N$ layers. The form of the Transformer depends on the input.

\noindent
\textit{Transformer for image-based inputs:} 
The Transformer attends to the temporal dimension in order to capture the temporal context. 
Sinusoidal positional encoding $P$ for the frame number (time) are added to the visual features $F_v$. The output is $T$ features of dimension $d$:

\begin{equation}
    E_v = TransformerEncoder_{img}(F_v + P) \in \mathbb{R}^{T\times d}
\end{equation}

\noindent
\textit{Transformer for vector-based inputs:}
For keypoint vector based inputs, we design a Factorised attention transformer~\cite{SpaceTimeAttn_2021}. The goal here is to enable the Transformer to attend to the keypoints as well as to the temporal context. Learnable positional embeddings are used to label both keypoints and time,  and are added to the visual features $F_v$. The Transformer has a stack of $N$ Encoder layers, each comprising a feed-forward layer and two attention blocks: the first attention block attends across the time dimension and the second block attends across the keypoints. The Transformer outputs $T$ features of dimension $N_{kp}\times d$:

\begin{equation}
    E_v = TransformerEncoder_{vec}(F_v + P) \in \mathbb{R}^{T\times N_{kp}\times d}
\end{equation}

\noindent
\textit{Aggregation:} We use a temporal aggregation layer which computes the mean for all the $T$-frames, followed by a linear layer. Thus, the output visual embedding $E_v$ is a $d$-dimensional vector for image inputs, and $N_{kp}$ $d$-dimensional vectors for keypoint vector inputs.

\subsection{Speech Representation}
For a speech segment $I_{speech}$, a mel-spectrogram representation is extracted using a window length of $25$ms with a hop length of $10$ms sampled at $16$kHz. The mel-spectrograms ($T', 80$) are given as input to the speech encoder which is similar to previous synchronisation networks~\cite{Afouras20b}. The encoder is a stack of $2$D convolutions with appropriate strides to match the visual time-steps $T$ and outputs $d$ dimension speech features $F_s$. We then use a linear layer to obtain speech embedding $E_s$. 

\subsection{Self-supervised Synchronisation Training}
\label{sec:synchronisation}
\label{sec:training}

For each visual embedding $E_{v}$, we ask how similar is this embedding to the corresponding
speech embedding $E_{s}$? To answer this, we compute the cosine similarity between the visual and speech embeddings. Note that for the keypoint vector input, the visual embedding also has the spatial keypoint dimension, thus we consider the maximum over the spatial response as is done in AVObjects~\cite{Afouras20b} to obtain the similarity.

\begin{figure}[h]
  \centering
\includegraphics[width=\linewidth]{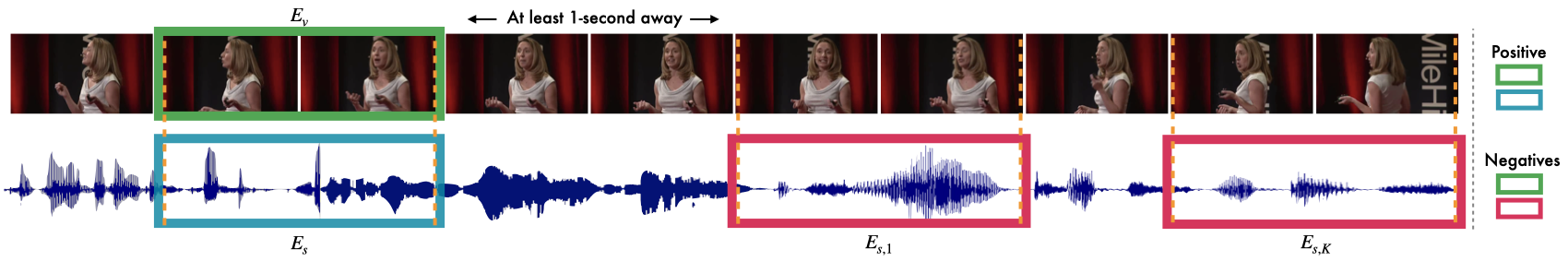}
  \vspace{-20pt}
  \caption{Illustration of self-supervised synchronisation training. The network is trained using contrastive loss by sampling $K$ negative speech segments which are at least $1$-second away from the positive segment.}
  \label{fig:sync_training}
  \vspace{-10pt}
\end{figure}

The model is trained for the task of audio-visual synchronisation: the objective is to maximize the similarity between the visual input and speech input when they are synchronized, and to minimize the similarity between the shifted versions of the audio input. We sample $K$ negative audio segments from the same video clip which are at least
$1$-second away from the positive window. This is illustrated in Figure~\ref{fig:sync_training}. Thus, given the visual embedding $E_{v}$, synchronized speech embedding $E_{s}$, and $K$ shifted speech embeddings $E_{s,1}, E_{s,2}, ..., E_{s,K}$, we minimize the contrastive loss:
\begin{equation}
    \mathcal{L} = -log \frac{exp(E_{v} \cdot E_{s})}{exp(E_{v} \cdot E_{s}) + \sum_{j=1}^{K}exp(E_{v} \cdot E_{s,j})}
\end{equation}

\subsection{Implementation Details}
\label{sec:implem}

\noindent
\textbf{Keypoint Extraction:}
To extract the keypoints, we utilize
Mediapipe~\cite{Mediapipe_2019_Lugaresi}, which is a publicly
available library built for human pose estimation. The major advantage
of using Mediapipe over other libraries, such as
OpenPose~\cite{OpenPose_2019_Cao}, is that it enables the keypoints to
be extracted in real-time (on a CPU), thus reducing the computational
burden. Mediapipe pose model generates $33$ pose keypoints, out of which we discard the (sparse) face and lip keypoints and use the remaining $22$ keypoints as inputs to the GestSync  model. This ensures that the model learns to focus on the gestures rather than face or lip movements. We render the keypoint skeleton images using the $22$ pose keypoint coordinates for the keypoint-image inputs, and use the normalized keypoint coordinates directly for the keypoint-vector inputs. For the task of lip synchronisation, we extract dense $128$ face keypoints using the Mediapipe face model, and these form the input to the lip synchronisation models.

\noindent
\textbf{Hyperparameters:} For training, videos are sampled at $25$ FPS, at a resolution of $270\times 480$ pixels. We extract $1$-second audio and video segments ($T=25$) which are considered as inputs. Our Transformer has $6$ encoder layers, with $8$ attention heads and generates $512$-dimension embeddings. We train our network using the Adam optimizer~\cite{AdamOptimizer_2014_Kingma} with a learning rate of $1e^{- 4}$. The training is stopped once the validation loss does not improve for $5$ epochs. 

\noindent
\textbf{Sampling:} The visual embeddings $E_{v,i}$ are computed for $T=25$ frames, at a stride of $1$ frame and the speech embeddings $E_{s,i}$ are computed for $T'=100$ frames at a stride of $4$ frames to obtain per-frame embeddings. The correlation between gestures and speech segments are not as fine-grained as that of lip synchronisation. Thus, for the task of gesture-sync: (i) we need longer context windows to enable the model learn well, and (ii) the negative speech segments need to be carefully sampled. It is possible that the gesture begins (ends) slightly before (after) the speech segment ends. To account for this, we ensure that the negatives are atleast $1$-second away from the positive window. 

\noindent
\textbf{Training procedure:} For keypoint vector based inputs, the visual embeddings also have a spatial dimension (keypoint dimension). To make the training easier, we start by taking the mean of the embeddings and then switch to the maximum as explained in Section~\ref{sec:synchronisation}. 

\vspace{-5pt}
\section{Experiments}
We start by describing the datasets and metrics used for our tasks in Section~\ref{sec:datasets}. Since there are no existing works on our task of gesture synchronisation, we start our evaluation on the standard lip synchronisation task and compare our different representations against the state-of-the-art work in Section~\ref{sec:lipsync}. We then move on to evaluate on the proposed gesture synchronisation task in Section~\ref{sec:handsync}. 

\vspace{-8pt}
\subsection{Datasets and Metrics}
\label{sec:datasets}

\noindent
\textbf{Datasets:}
We conduct our experiments on the LRS3 dataset~\cite{LRS3_2018_Afouras}, which consists of TED videos downloaded from YouTube, totaling over $400$ hours of video data. The dataset features a diverse set of speakers (over $9$K) and a large vocabulary. We use the full-frame videos and resize them to ($270\times 480$) pixels. To ensure a fair evaluation, we create a new test split from the LRS3 pre-train set, as the videos in the original LRS3 test set are short in duration. Our train and test sets do not have any speaker overlaps, and we have released our code, models, and data splits to encourage future research. For the task of lip-synchronisation, we report the results on LRS2 test set~\cite{Afouras19} to enable a direct comparison with the existing works.\\

\noindent
\textbf{Metrics:}
For evaluating lip synchronisation, we follow the same protocol as previous lip-sync methods~\cite{Chung2019perfect, Afouras20b}. The video is shifted within $\pm15$ frames ($31$ negatives in total), and the synchronisation is considered correct if the predicted offset is within $1$ video frame of the ground truth offset. We obtain per-frame scores by computing the cosine similarity of the visual and speech embeddings. The scores are averaged over $F$ input frames (here $F=[5, 7, 9, 11, 13, 15]$). A random prediction would give an accuracy of $9.7\%$ ($3$ out of $31$ are correct). Lip movements rapidly change and are very fine-grained, meaning that there is a significant difference between consecutive frames. On the other hand, this is not the case for gestures. Thus, to evaluate gesture synchronisation, we shift the video by $\pm50$ frames with a gap of $5$ frames between each negative ($21$ negatives in total) and average the per-frame score over $F'$ input frames (here $F' = [25, 50, 75, 100]$). We deem the synchronisation to be correct if the predicted offset is within $\pm10$ frames of the ground truth. Note that since gestures and speech are coarsely correlated, we cannot expect the model to predict the finer-grained shifts (i.e.\ shifts less than $\pm10$ frames). Using this setup yields a chance accuracy of $23.8\%$ (any of the closest $5$ shifted versions out of the $21$ is correct). 

\subsection{Lip Synchronisation}
\label{sec:lipsync}

\noindent
\textbf{Comparison:} We train our keypoint representation based models, both keypoint-image and keypoint-vector based methods for the task of lip synchronisation. We use the same settings as done in previous works, where the temporal window of $5$ frames is considered for training and testing. We compare our models with three previous models: (i) SyncNet~\cite{Chung16a}, (ii) Perfect Match~\cite{Chung2019perfect}, and (iii) AVObjects~\cite{Afouras20b}. We start by re-implementing the existing RGB based AVObjects model, followed by our keypoint based models as explained in Section~\ref{sec:arch}. Our keypoint based models use $128$ face keypoints.

\noindent
\textbf{Results:} Table~\ref{table:lipsync} presents the lip synchronisation results for different number of frames used for averaging the scores on the LRS2 dataset~\cite{Afouras19}. Our RGB model's scores validates the correctness of our experimental setup. We can observe that there is no significant difference between our keypoint based methods and the existing models. The fact that we are able to achieve an accuracy as high as $\approx 95\%$ using keypoint representations is remarkable and demonstrates the effectiveness of using the keypoint representations. 

\begin{table}[ht]
    \centering
    \setlength{\tabcolsep}{5pt}
    \caption{Performance comparison of lip synchronisation accuracy ($\%$) averaged over a given number of input frames on LRS2 dataset~\cite{Afouras19}. 
    Our keypoint based methods achieve very similar results to the existing works, but with an advantage of being computationally efficient.}
    
    \begin{tabular}{c|cccccc|c}
    \hline

    \textbf{Method} & \textbf{5} & \textbf{7} & \textbf{9} & \textbf{11} & \textbf{13} & \textbf{15} & \textbf{Input type}\\
    \hline

    SyncNet~\cite{Chung16a} & 75.8 & 82.3 & 87.6 & 91.8 & 94.5 & 96.1 & face crop\\
    PerfectMatch~\cite{Chung2019perfect} & 88.1 & 93.8 & 96.4 & 97.9 & 98.7 & 99.1 & face crop\\
    AVObjects~\cite{Afouras20b} & 78.8 & 87.1 & 92.1 & 94.8 & 96.3 & 97.3 & full image\\
    \hline
    RGB (AVObjects re-imp.) & 80.2 & 88.1 & 92.0 & 95.5 & 96.1 & 97.5 & full image\\
    Ours: Keypoint-image & 69.7 & 76.2 & 83.4 & 86.9 & 89.9 & 91.3 & face points\\
    Ours: Keypoint-vector  & 71.9 & 78.8 & 84.3 & 89.5 & 92.0 & 94.9 & face points\\

    \hline
    \end{tabular}
    \vspace{-10pt}
    \label{table:lipsync}
\end{table}

\subsection{Gesture Synchronisation}
\label{sec:handsync}

\noindent
\textbf{Baseline:} As there are no existing works on our proposed task, we implement an RGB baseline model similar to AVObjects~\cite{Afouras20b}, but with a longer temporal window ($25$ frames) and an LSTM layer to aggregate the temporal information (Conv.\ + LSTM model). Note for this model and others that use RGB frames as input, the face is occluded by overlaying a mask so that the lips, and face/head movements cannot be used for the task.

\noindent
\textbf{Comparison:} We present results on the Transformer network as explained in Section~\ref{sec:arch}. We train the same models using keypoint-image representation. For the keypoint vector representation, we experiment with a baseline Conv.\ + LSTM model, followed by a Transformer model where we attend to the keypoint dimension, and finally show results on the Transformer with factorised attention. All our keypoint based models utilize $22$ pose keypoints.

\noindent
\textbf{Results:} Table~\ref{table:handsync} gives the results for the task of gesture synchronisation on the LRS3 dataset. 
The difference in scores between the lip synchronisation and gesture synchronisation validates that gesture-sync is a far more challenging problem. We can observe that for this task, the raw RGB based Transformer model performs the best. In all the three representations, the Transformer based models outperform the convolutional baselines. Unlike the task of lip-synchronisation, here there is a gap in performance between the keypoint based representations and the RGB representation. We speculate that one of the reasons for this difference could be that face or lips do not need any $3$D information. However, this is not the case with gestures, as there could be occlusions present (for example, one arm in front of the other) and $3$D motion (e.g.\ movements towards the camera). This information is completely lost when we consider $2$D keypoint representations. Since the images naturally have this information available, their performance is superior when compared to the keypoints.

\vspace{-5pt}
\begin{table}[ht]
    \centering
    \caption{Performance comparison of gesture synchronisation accuracy ($\%$) averaged over a given number of input frames on the LRS3 dataset~\cite{LRS3_2018_Afouras}. 
    For the task of gesture-sync, RGB based model gives the best performance. }

    \begin{tabular}{c|c|cccc}
    \hline

    \textbf{Input} & \textbf{Method} & \textbf{25} & \textbf{50} & \textbf{75} & \textbf{100}\\
    \hline

    \multirow{2}{*}{RGB} & Conv.\ + LSTM baseline & 49.8 & 59.3 & 68.2 & 75.7\\
    & \textbf{Ours (Image based Transformer)} & 53.7 & 66.1 & 72.6 & 77.5\\
    \hline
    
    \multirow{2}{*}{Keypoint-image} & Conv.\ + LSTM baseline &  39.1 & 47.2 & 52.3 & 63.6\\
    & \textbf{Ours (Image based Transformer)} & 43.0 & 51.0 & 58.2 & 61.6\\
    \hline
    
    \multirow{3}{*}{Keypoint-vector} & Conv.\ + LSTM baseline & 38.2 & 40.3 & 47.7 & 54.8 \\
    & Transformer (with temporal attn.) & 40.6 & 43.5 & 49.9 & 56.5 \\
    & \textbf{Ours (Transformer with Fact. attn.)} & 41.7 & 49.8 & 58.1 & 62.7\\

    \hline
    \end{tabular}
    \vspace{-5pt}
    \label{table:handsync}
\end{table}

\begin{figure}[h]
  \centering
\includegraphics[width=\linewidth]{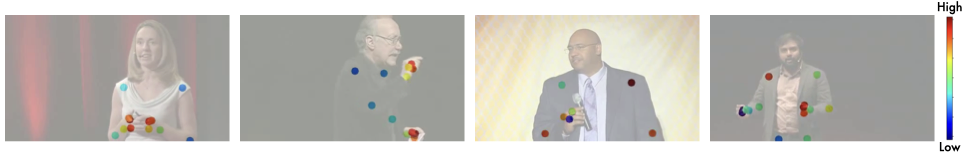}
  \vspace{-20pt}
  \caption{Attention map visualization of the keypoint-vector Transformer model (with factorized attention). We can observe that the model focuses mostly on hands/arms to identify if the gestures are in-sync with speech. Note that the actual speaker frames are overlayed only for visualization.}
  \label{fig:visualization}
  \vspace{-10pt}
\end{figure}

\noindent
\textbf{Visualization:}
Figure~\ref{fig:visualization} depicts the visualization of the attention map from the keypoint-vector based model. The attention map indicates that the model is highly attentive to hands and arms amongst all the keypoints. 

\begin{wraptable}{r}{6.4cm}
    \vspace{-12pt}
    \small
    \setlength{\tabcolsep}{2.5pt}
    \caption{Comparison of the model size (in million parameters) and the inference time (in seconds). The keypoint-vector based model has $\approx 1.4 \times$ fewer parameters and is $\approx 6 \times$ faster than the best performing RGB based Transformer model.}
    \begin{tabular}{c|cccc}
    \hline

    & RGB & Kp-img & Kp-vec \\
    \hline

    $\#$ params (M)$\downarrow$ &  29.2 & 29.2 & \textbf{21.4} \\    
    inf. time (sec)$\downarrow$ & 12.12 & 10.81 & \textbf{2.20} \\

    \hline
    \end{tabular}
    \label{table:params}
    \vspace{-10pt}
\end{wraptable} 

\noindent
\textbf{Computation Comparison:}
In Table~\ref{table:params}, we compare the model parameters and the inference time for the best performing models in each of the input representations. For comparing the inference time, we report the time taken to load and generate embeddings for a $5$-second input on a single NVIDIA V$100$ GPU. As we can see from the table, the keypoint-vector based model is very efficient: it is $5.5\times$ and $4.9\times$ faster than raw RGB and keypoint skeleton based models respectively. Thus, in terms of compute, the keypoint-vector based model beats other models by a large margin. Note that we discard the time needed for keypoint extraction here for all the models, since Mediapipe runs in real-time on a CPU.

\vspace*{-10pt}
\subsubsection{Bridging the gap between RGB and keypoint-vector representations}

One of the major advantages of using the keypoint representations is the \textit{computational efficiency}: keypoint-vector representations are $\approx6\times$ faster than RGB representations. However, as observed in Table~\ref{table:handsync}, this comes with a cost in the performance. Naturally a question arises is there a possibility to close or reduce the gap that exists between the RGB and keypoint representations? Below, we describe the techniques we experimented to achieve this feat.

In comparison to the RGB model which can \textit{see} the entire frame -- the arms of the person, not just their joints for example -- the keypoint model has much less information. Thus, to provide more information, we gradually increase the number of keypoints between the joints, starting from $22$ to extremely dense representations up to $142$ keypoints. The results are presented in Table~\ref{table:dense_kps}. We can observe that the $142$ keypoints model achieved a significant boost in performance, reducing the gap between the RGB and keypoint representations from $14.4\%$ to $7.9\%$. However, there is a trade-off in terms of computational complexity (calculated for a $5$-second input on a single NVIDIA V$100$ GPU), as indicated in Table~\ref{table:dense_kps} (last column). Improving the performance of the keypoint-vector representation, while keeping the computational cost as low as possible is an interesting future direction to explore. Please refer to our Appendix for additional experiments and ablation studies.

\begin{table}[ht]
    \centering
    \setlength{\tabcolsep}{6pt}
    \caption{Performance comparison of gesture synchronisation accuracy ($\%$) with different number of keypoints averaged over a given number of input frames ($25$, $50$, $75$, $100$) on the LRS3 dataset~\cite{LRS3_2018_Afouras}. The dense keypoint model with $142$ keypoints gives the best performance among other keypoint models.}
    
    \begin{tabular}{c|c|cccc|c}
    \hline

    \textbf{Input} & \textbf{\# keypoints} & \textbf{25} & \textbf{50} & \textbf{75} & \textbf{100} & \textbf{inf. time (sec)} $\downarrow$\\
    \hline

    Keypoint-vector & 22 & 41.7 & 49.8 & 58.1 & 62.7 & \textbf{2.20}\\
    Keypoint-vector & 48 & 44.7 & 52.7 & 60.2 & 67.0 & 4.32\\
    Keypoint-vector & 70 & 45.8 & 55.6 & 62.3 & 67.2 & 6.27\\
    Keypoint-vector & 142 & \textbf{46.1} & \textbf{57.5} & \textbf{65.6} & \textbf{68.9} & 12.10\\
    \hline
    RGB & - & 53.7 & 66.1 & 72.6 & 77.5 & 12.12\\

    \hline
    \end{tabular}
    \vspace{-15pt}
    \label{table:dense_kps}
\end{table}

\section{Application: Who is speaking?}

We investigate our model's ability to identify the speaker when multiple individuals are present in the scene (as in Figure~\ref{fig:teaser}). The goal is to recognize the target speaker using our trained model, without looking at their faces. There could be numerous scenarios where the face, or more precisely the lip movements of the speaker is corrupted, occluded, not covered by the camera angle or not present at all. In such cases, the traditional methods of spotting the speakers using lip-synchronisation fails, whereas our model can be used as an  alternative. 

\noindent
\textbf{Procedure and Evaluation:}
For this experiment, we create a pairwise test set by mixing the video of the target speaker with videos of other randomly selected speakers from the LRS3 dataset. Next, we calculate the similarity between the visual embedding of the target speaker and the speech embeddings of all the speakers to determine the likelihood of the speech belonging to the target speaker. Table~\ref{table:applications} presents the model's performance on the LRS3 dataset, where we report the accuracy of correctly identifying the target speaker among $1$, $3$, and $5$ negative speakers. A random prediction for each of these settings would result in $50\%$, $25\%$, and $16\%$ accuracy, respectively.

\noindent
\textbf{Results:}
We report the scores for the top-performing models (selected from Table~\ref{table:handsync}) for each of the representations. The model's performance is remarkably impressive, particularly in the case of $5$ negative speakers, where our RGB-based model achieves an accuracy of $73.2\%$. This suggests a significant correlation between the speaker's body movements and speech, which our model captures well. It also indicates the high quality of our learned visual and speech embeddings.

\begin{table}[ht]
    \centering
    \caption{Performance evaluation of identifying the target speaker in a crowd on the LRS3 dataset~\cite{LRS3_2018_Afouras}. The accuracy of determining the positive speaker in the presence of: (i) 1-negative, (ii) 3-negative, and (iii) 5-negative speakers is computed.}

    \begin{tabular}{c|ccc}
    \hline

    \textbf{Method} & \textbf{1-negative} &\textbf{3-negatives} & \textbf{5-negatives} \\
    \hline

    Ours - RGB & 87.0 & 77.6 & 73.2 \\
    Ours - Keypoint-image & 82.4 & 70.6 & 62.6 \\
    Ours - Keypoint-vector & 79.0 & 58.7 & 50.1\\

    \hline
    \end{tabular}
    \vspace{-10pt}
    \label{table:applications}
\end{table}

\section{Summary, limitations, and future directions}

In this work, we have presented a new task of gesture-speech synchronisation, which has not been explored before. We provided a detailed analysis on the different types of input representations, including raw RGB frames, keypoint skeleton images, and keypoint vectors. We demonstrated that it is indeed possible to identify ``who is speaking'' in a crowd using our model, by focusing only on the gestures without a need for the `talking head' to be visible. 

Compared to using lips to identify the speaker in a crowd, gestures suffer from the problem that some people do not gesture extensively. Also, a longer temporal context is required for gesture-sync compared to lip-sync. 

There are several interesting directions to explore in the future, such as the correlation between gestures and language. In \cite{GesturesLanguage_2022_ICCL}, initial results on predicting the native language of the speakers using gestures has been provided. Another direction is to explore if it is possible for some types of gestures (such as iconic or metaphoric) to `read' the semantics of what is being said.

\vspace{2mm}
\noindent
\textbf{Acknowledgements.}
The authors would like to thank K R Prajwal, Tengda Han, Triantafyllos Afouras and Joon Son Chung  for helpful discussions. This research is funded by  EPSRC Programme Grant VisualAI EP/T028572/1, and a Royal Society Research Professorship RP \textbackslash R1 \textbackslash 191132.

\bibliography{shortstrings,vgg_local,extra,egbib}

\newpage
\appendix

\section{Ablation Studies}
We perform ablation experiments using our best keypoint-vector based model. The results are for the LRS3 {\em validation} set~\cite{LRS3_2018_Afouras}, and are used to choose the best model parameters. The results in the main paper are for the LRS3 {\em test} set.

\subsection{Temporal window}
We provide justification for using a temporal window of $T=25$ frames ($1$-second of audio and video segments) as input to our model in Table~\ref{table:abl_temporal_window}. Note that $T$ specifies the input to the model, whereas $F=[25,50,75,100]$ specifies the number of frames used when averaging the audio-visual similarity score. Our model consistently improves in performance as the length of the input segment increases, which aligns with our intuition that longer context windows are beneficial due to the sparse correlation between gestures and speech. From Table~\ref{table:abl_temporal_window}, it is evident that using only $5$ frames leads to a significant deterioration in performance. When using a longer window of $T=50$ frames, the performance is similar to that of $T=25$ frames, with only a minor improvement. This behavior can be attributed to the limited availability of training data for longer input segments. Since we utilize a contrastive loss framework where shifted versions of the same video act as negative samples, obtaining sufficiently long videos for sampling negatives becomes challenging. Moreover, as explained in Section $3.4$ of the main paper, our negative samples need to be at least 1 second away from the positive sample, further limiting the sampling process for longer videos when training the model with $T \geq 50$ frames.

\begin{table}[ht]
    \centering
    \caption{Synchronization performance variation on LRS3 val. The first column specifies the input window (the number of frames input to the GestureSync network). The variation across the columns specifies the number of frames used to average the score. Video and audio are sampled at 25 Hz. Longer input windows enable the model to capture the temporal context and effectively learn the gesture-speech synchronisation.}

    \begin{tabular}{c|cccc}
    \hline

    \textbf{Temporal Window} & \textbf{25} & \textbf{50} & \textbf{75} & \textbf{100} \\
    \hline

    5 & 32.1 & 39.5 & 46.2 & 51.7\\
    15 & 40.4 & 48.9 & 57.6 & 63.2\\
    25 & 43.2 & 51.5 & 58.8 & 64.1\\
    50 & 44.5 & 51.7 & 58.9 & 64.3\\
 
    \hline
    \end{tabular}
    \label{table:abl_temporal_window}
\end{table}

\subsection{Using additional hand keypoints}
\label{sec:ablation_kps}

We show the effect of using additional hand keypoints in Table~\ref{table:abl_additioanl_kps}. Specifically, we utilize the hand keypoints (extracted from Mediapipe~\cite{Mediapipe_2019_Lugaresi}) along with the pose keypoints. This gives a total of $64$ keypoints which act as input to the GestureSync model ($22$ pose keypoints + $21$ keypoints for each hand). Using more keypoints gives us an improvement across all the averaging windows $F$, except for the largest window of $100$ frames.

\begin{table}[ht]
    \centering
    \caption{Comparison of using additional hand keypoints along with pose keypoints. Synchronization performance variation on LRS3 val. }

    \begin{tabular}{c|cccc}
    \hline

    \textbf{Method} & \textbf{25} & \textbf{50} & \textbf{75} & \textbf{100}\\
    \hline

    pose (22 kps) & 43.2 & 51.5 & 58.8 & 64.1\\
    pose + hands (64 kps) & 45.3 & 52.9 & 59.0 & 62.2\\
 
    \hline
    \end{tabular}
    \label{table:abl_additioanl_kps}
\end{table}

\section{Additional Experiments}

\subsection{Evaluation on LRS3-Lang dataset}

In addition to the evaluation on the LRS3 dataset~\cite{LRS3_2018_Afouras} shown in the main paper (Section $4.3$), we also assess the model on the  LRS3-lang dataset~\cite{Afouras20c}. LRS3-lang is a multi-lingual dataset comprising $12$ different languages with a total of over $1300$ hours of video data. We obtained this dataset for our evaluation from the authors (since the data has not yet been publicly released). Note that the pre-processing used in the provided body-crops data (including resolution of videos, and the bounding-box used to create the body crops) is entirely different from that of LRS3. Thus, we fine-tune our models (for $5$ epochs) to adapt to the different pre-processing settings. Since the official train-test splits are not yet provided for this dataset, we randomly sample $\sim 3\%$ videos without any speaker overlaps to create our test set (we will release the splits). The distribution of train and test sets across multiple languages is shown in Figure~\ref{fig:lrs3_lang_distribution}. The evaluation on a more challenging, multi-lingual LRS3-lang dataset highlights the capabilities of the GestureSync model and shows that it is not limited by language barriers.

Table~\ref{table:handsync_lrs3lang} shows the results of the models using different input representations on the LRS3-lang dataset~\cite{Afouras20c}. In-line with the LRS3 results shown in the main paper (Table $2$), our RGB-based Transformer model achieves the best performance. It is worth noting that the synchronisation accuracy on LRS3-lang is very similar to that of LRS3, despite the fact that LRS3-lang is a much harder dataset with a wider variations in terms of speakers, languages, and resolution. This indicates the abilities of our model to work effectively in challenging settings. 
inputs.

\begin{table}[ht]
    \centering
    \caption{Performance comparison of gesture synchronisation accuracy ($\%$) averaged over a given number of frames $F$ on the LRS3-lang dataset~\cite{Afouras20c}.}

    \begin{tabular}{c|cccc}
    \hline

    \textbf{Method} & \textbf{25} & \textbf{50} & \textbf{75} & \textbf{100}\\
    \hline

    Ours - RGB & 44.8 & 56.0 & 67.9 & 76.4\\
    \hline
    
    Ours - Keypoint-image & 37.2 & 47.4 & 52.3 & 58.9\\
    \hline
    
    Ours - Keypoint-vector & 40.1 & 46.7 & 54.5 & 62.5\\

    \hline
    \end{tabular}
    \label{table:handsync_lrs3lang}
\end{table}

\subsection{Do gestures vary with the language spoken?}

\begin{figure}[ht]
\includegraphics[width=\textwidth]{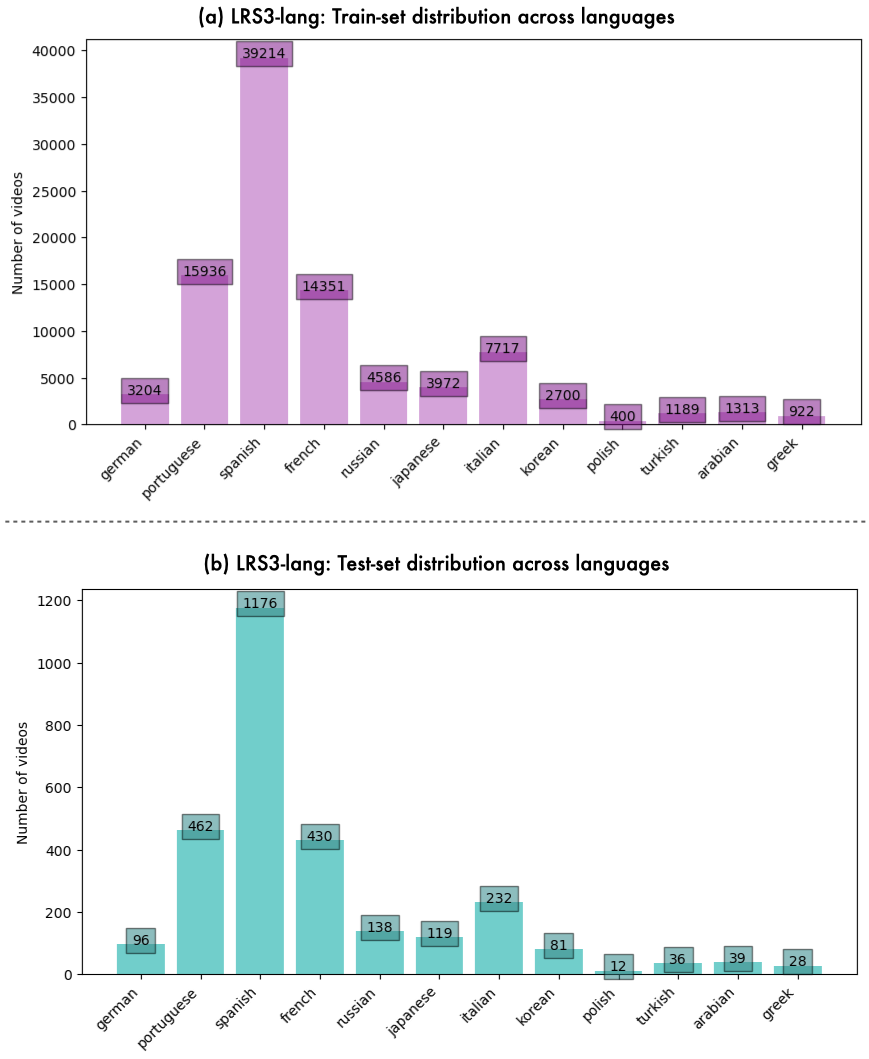}
  \caption{We show the distribution of languages in the training and test sets of LRS3-lang dataset~\cite{Afouras20c}.}
  \label{fig:lrs3_lang_distribution}
\end{figure}

We investigate the variations in gesture-speech correlations across different languages (and hence to some extent nationalities) worldwide. To conduct this analysis, we utilize the LRS3-lang dataset, and leverage the provided language labels to categorize the speakers accordingly. We specifically evaluate on eight languages, excluding Polish, Turkish, Arabian and Greek due to the limited availability of test data in these languages (see Figure~\ref{fig:lrs3_lang_distribution} (b)). 

In Table~\ref{table:handsync_lrs3lang}, we compute the synchronisation accuracy (averaged over $F$ frames, where $F=[25, 50, 75, 100]$) for each language category individually. It is evident that certain languages exhibit a stronger and more explicit correlation, such as Italian, Portuguese, Spanish, and French. Conversely, the correlation between gestures and speech is less pronounced for Korean and Japanese speakers, showcasing an opposite trend.

\begin{table}[ht]
    \centering
    \caption{We study how the synchronization performance of the gesture-speech model varies across different languages using the LRS3-lang dataset. While some languages like Italian, Portuguese, Spanish have stronger gesture-speech correlations, languages like Japanese and Korean have weaker links.}

    \begin{tabular}{c|cccc}
    \hline

    \textbf{Language} & \textbf{25} & \textbf{50} & \textbf{75} & \textbf{100}\\
    \hline

    german & 40.9 & 45.5 & 68.2 & 75.0\\
    portuguese & 44.9 & 60.5 & 71.5 & 73.8\\
    spanish & 46.8 & 59.5 & 72.2 & 80.3\\
    french & 47.1 & 58.9 & 65.5 & 78.2\\
    russian & 42.2 & 54.2 & 63.1 & 72.0\\
    japanese & 39.7 & 42.8 & 55.6 & 66.7\\
    italian & \textbf{49.3} & \textbf{61.6} & \textbf{74.0} & \textbf{82.3}\\
    korean & 35.1 & 38.9 & 49.6 & 64.9\\

    \hline
    \end{tabular}
    \label{table:ablation_nationalities}
\end{table}

\subsection{Sensitivity to speaker attributes}
We analyse the behaviour of the GestureSync model on the different speaker attributes such as gender and age of the speakers. For gender classification and age estimation, we use OpenCV-based public implementation\footnote{\url{https://github.com/smahesh29/Gender-and-Age-Detection}} and obtain the labels on the LRS3 test set~\cite{LRS3_2018_Afouras}. 

Table~\ref{table:speaker_attributes} demonstrates that the synchronization performance is higher in the female category compared to the male category. This observation suggests that there may be inherent differences in the gestures and speech patterns of male and female speakers. Further investigation and analysis of these gender-related differences could provide valuable insights into the dynamics of correlation between gestures and speech. Table~\ref{table:speaker_attributes} also reveals that there is no significant difference in performance across different age groups. This observation could be attributed to the nature of the LRS3 dataset, which predominantly consists of videos from trained TED speakers. The dataset's composition of experienced speakers might mitigate any potential impact of age on the performance of speech-gesture synchronization. Therefore, the age of the speakers does not appear to be a significant contributing factor in determining the level of synchronization achieved in this context.

\begin{table}[ht]
    \centering
    \caption{Effect of the speaker attributes such as gender and age on model's synchronization performance on LRS3 test set~\cite{LRS3_2018_Afouras}. The GestureSync network performs better for female category, whereas the performance remains consistent across different age-groups.}
    \begin{tabular}{lccccc}
    \hline
 
    \textbf{Attribute} & \textbf{Class} & \textbf{25} & \textbf{50} & \textbf{75} & \textbf{100} \\
    \hline
    
    \multirow{2}{*}{Gender} & Female & 42.9 & 50.5 & 59.2 & 64.4\\
    & Male & 41.1 & 48.2 & 56.2 & 61.2\\
    
    \hline
    
    \multirow{3}{*}{Age} & $<25$ & 42.8 & 48.0 & 56.3 & 62.9\\
    & $25-60$ & 42.1 & 49.4 & 56.1 & 62.4\\
    & $>60$ & 42.9 & 49.1 & 57.1 & 63.2\\
    
    \hline

    \end{tabular}
    \vspace{-10pt}
    \label{table:speaker_attributes}
\end{table}

\subsection{Improving the keypoint-vector representation}

As explained in Section $4.3.1$ of the paper, one of our long-term goals is to bridge the gap between RGB and keypoint-vector representations. We perform several further experiments as explained below to investigate to what extent we can boost the keypoint-vector model's performance.

\subsubsection{Data Augmentation}
One of the techniques which has proven to be beneficial in various image and video processing tasks is data augmentation. Strategies such as translation, flipping, rotation, and scaling are used to improve the robustness of the model, thus enhancing the performance during inference. Following these traditional techniques, we too apply data augmentation to our keypoint-vector representation network. Specifically, we apply the following augmentations: (i) Shifting -- Randomly shift the $x$ and $y$ co-ordinates of the keypoints in the range of $[-50, 50]$, (ii) Rotation -- Rotate all the keypoints by an angle in the range of $[-10, 10]$, (iii) Scaling -- Scale the keypoints randomly in the range of $[0.7, 1.3]$. Table~\ref{table:data_augmentation} shows the results of augmentations. We can observe that using data augmentation techniques results in further boosts in the performance.

\begin{table}[ht]
    \centering
    \caption{Comparison of adding data augmentation to keypoint-vector representation model on LRS3 test set. Adding augmentation helps in improving the performance.}

    \begin{tabular}{c|cccc}
    \hline

    \textbf{Method} & \textbf{25} & \textbf{50} & \textbf{75} & \textbf{100}\\
    \hline
    
    W/o augmentation & 41.7 & 49.8 & 58.1 & 62.7\\
    With augmentation & 43.1 & 51.2 & 59.5 & 64.2\\
    
    \hline
    \end{tabular}
    \vspace{-10pt}
    \label{table:data_augmentation}
\end{table}

\subsubsection{Using head motion information}

When the person is speaking, head movements convey vital natural motion information alongside gestures. While we opt $not$ to incorporate lip motion data in our study, we have the flexibility to make use of head motion when it is accessible. We extract the face keypoints using Mediapipe and consider the face-oval/head keypoints along with the pose keypoints for this experiment. Utilizing all face-oval keypoints could potentially introduce lip motion information from the lower jaw regions, so to completely avoid any lip-related input, we conduct another experiment focusing solely on the upper head (above the ears). The outcomes are presented in Table~\ref{table:head_motion}. Remarkably, we achieve an almost perfect score of $95.7\%$ with a $100$-frame average when utilizing all head keypoints. As previously mentioned, this performance can be attributed to potential lip motion leakage. Notably, the performance of pose combined with upper head keypoints (the last row in the table) demonstrates a substantial enhancement compared to using only pose keypoints, highlighting the significant role of head motion in determining synchronisation.

\begin{table}[ht]
    \centering
    \caption{Performance comparison of utilizing the head motion information in determining synchronisation (on LRS3 test set).}

    \begin{tabular}{c|cccc}
    \hline

    \textbf{Method} & \textbf{25} & \textbf{50} & \textbf{75} & \textbf{100}\\
    \hline
    
    pose (22 kps) & 41.7 & 49.8 & 58.1 & 62.7\\
    pose + head (58 kps) & 77.6 & 88.6 & 94.3 & 95.7\\
    pose + upper head (43 kps) & 49.8 & 60.9 & 70.1 & 76.2\\
    \hline
    \end{tabular}
    \label{table:head_motion}
\end{table}

\section{Visualisation}

\subsection{RGB input representation: Masked frames}

\begin{figure}[ht]
\includegraphics[width=\textwidth]{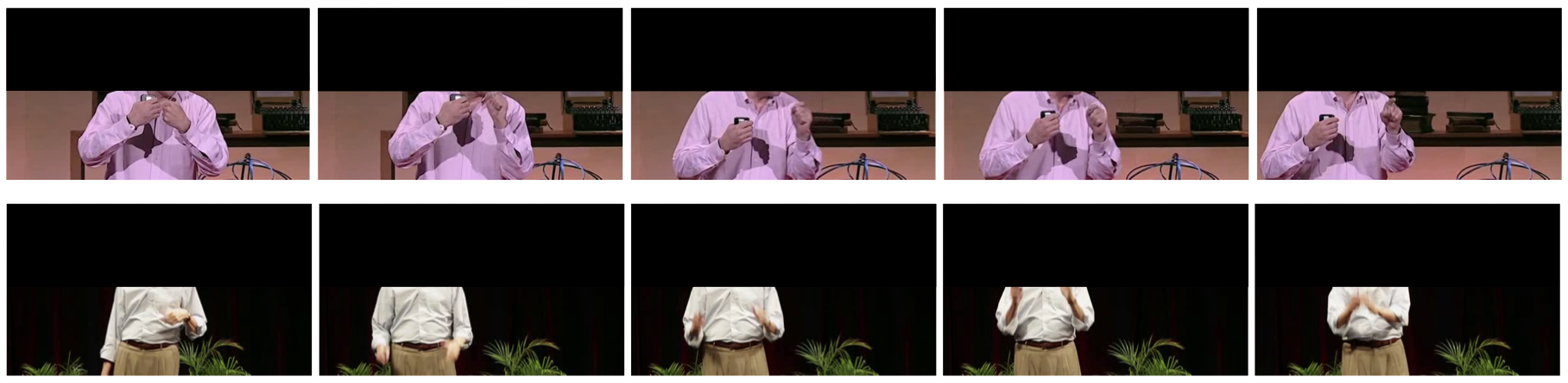}
  \caption{Visualisation of input frames for RGB representation based network for gesture synchronisation. The face is masked to avoid using any lip motion information.}
  \label{fig:rgb_masked_samples}
\end{figure}

Figure~\ref{fig:rgb_masked_samples} demonstrates the input masked frames for our RGB representation based model. To avoid using face and lip motion information, we mask the face region as shown in the figure.

\subsection{Gesture variation} 

\begin{figure}[ht]
\includegraphics[width=\textwidth]{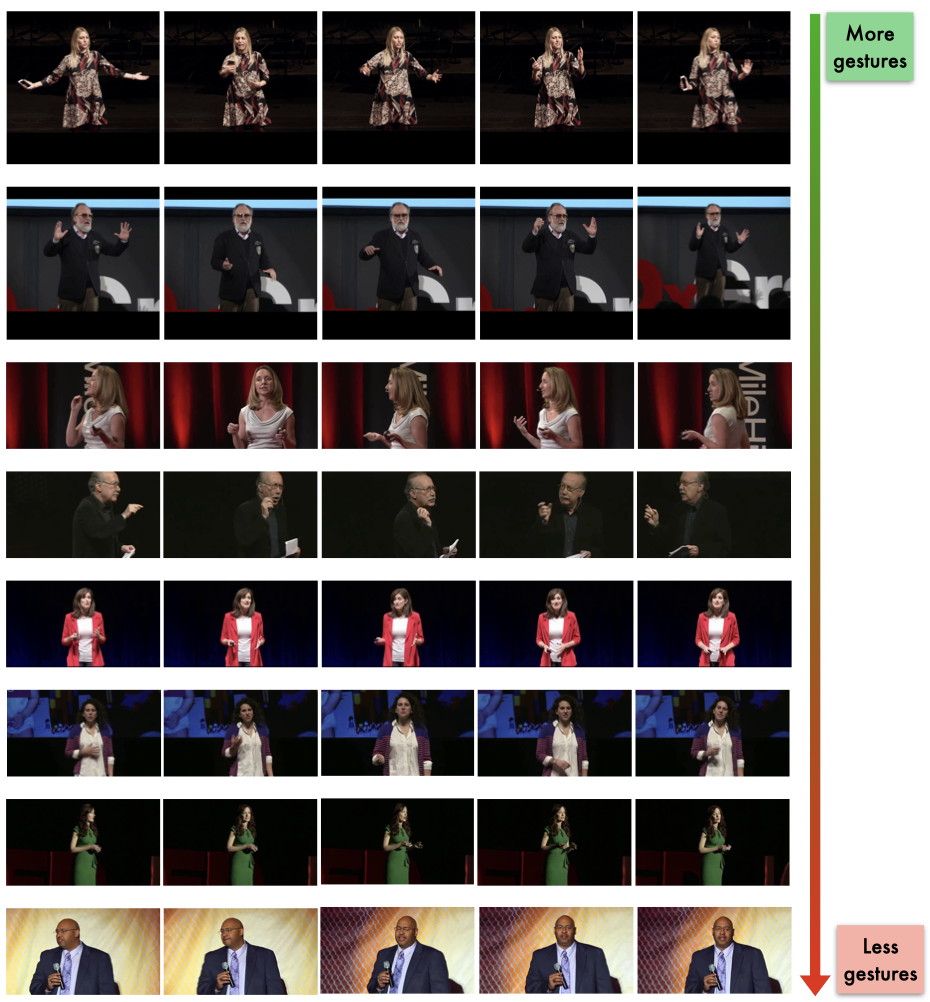}
  \caption{Variation of gestures across different speakers. A few speakers exhibit clear and expressive gestures (top rows) while others exhibit less prominent gestures (bottom rows). This diversity highlights the inherent challenges associated with our task.}
  \label{fig:handsync_samples}
\end{figure}

Figure~\ref{fig:handsync_samples} illustrates the variation of gestures across different speakers, highlighting that not all speakers exhibit expressive gestures that can be readily associated with their speech. These variations pose a challenge for our task. However, by providing a longer temporal context, the model can effectively aggregate subtle cues and make confident predictions regarding the synchronization between speech and gesture. 

\end{document}